\documentclass[letterpaper, 10 pt, conference]{ieeeconf}  %

\IEEEoverridecommandlockouts                              %

\usepackage{mathptmx} %
\usepackage{amsmath} %
\usepackage{amssymb}  %

\usepackage{graphicx}
\usepackage{color}
\usepackage{array} %
\usepackage{multirow}
\usepackage{hyperref}
\usepackage{subcaption}
\usepackage{algorithm,algpseudocode}
\usepackage{xr}

\newcommand{\thisWork}{IVOA}
\newcommand{\mypara}[1]{{\smallskip\noindent \bf #1.}\hspace{0.1in}}

\newcolumntype{M}[1]{>{\centering\arraybackslash}m{#1}}

\algnewcommand\algorithmicforeach{\textbf{for each}}
\algdef{S}[FOR]{ForEach}[1]{\algorithmicforeach\ #1\ \algorithmicdo}

\title{\LARGE \bf
IVOA: Introspective Vision for Obstacle Avoidance
}

\author{Sadegh Rabiee$^{1}$ and Joydeep Biswas$^{1}$%
\thanks{$^{1}$Sadegh Rabiee and Joydeep Biswas are with the College of Information and Computer Sciences,
        University of Massachusetts Amherst, Amherst, MA, USA. Email:
        {\tt\small \{srabiee, joydeepb\}@cs.umass.edu }}
}

\begin{document}

\maketitle
\thispagestyle{empty}
\pagestyle{empty}

\begin{abstract}
Vision, as an inexpensive yet information rich sensor, is commonly used for perception on autonomous mobile robots. 
Unfortunately, accurate vision-based perception requires a number of assumptions about the environment to hold -- some examples of such assumptions, depending on the perception algorithm at hand, include purely lambertian surfaces, texture-rich scenes, absence of aliasing features, and refractive surfaces. In this paper, we present an approach for \emph{introspective vision} for obstacle avoidance (IVOA) -- by leveraging a supervisory sensor that is occasionally available, we detect failures of stereo vision-based perception from divergence in plans generated by vision and the supervisory sensor. By projecting the 3D coordinates where the plans agree and disagree onto the images used for vision-based perception, IVOA generates a training set of reliable and unreliable image patches for perception. We then use this training dataset to learn a model of which image patches are likely to cause failures of the vision-based perception algorithm. Using this model, IVOA is then able to predict whether the relevant image patches in the observed images are likely to cause failures due to vision (both false positives and false negatives). We empirically demonstrate with extensive real-world data from both indoor and outdoor environments, the ability of IVOA to accurately predict the failures of two distinct vision algorithms.
\end{abstract}

\section{INTRODUCTION} \label{introduction}

Recent developments in 3D LiDAR technology have facilitated robust obstacle avoidance for mobile robots and autonomous vehicles. Yet, the high cost of such sensors makes them inaccessible for a wide range of robotic applications, leaving vision as the next best option. Vision systems, on the other hand, are prone to errors from various sources such as image saturation, blur, texture-less scenes, etc. This has motivated the question of whether we can develop competency-aware vision systems capable of predicting their own failures and estimating their level of uncertainty. 

In this work we present an approach for introspective vision for obstacle avoidance (IVOA). The main idea behind IVOA is to equip one robot with a high-fidelity depth sensor and let the vision system use that as ground truth for learning of an introspection model that predicts failures of the stereo obstacle detection system. The introspection model can then be transferred to other agents, empowering them to predict failure cases of the vision-based obstacle detection, without having depth sensors for each robot. IVOA applies to any stereo vision-based obstacle detection system and provides the means for \textbf{self-supervised} training of an introspection model that predicts the probability of different types of failure (false positive and false negative) and pinpoints the location of the error on the input image. The proposed approach also provides a measure of uncertainty for its predictions. The benefits of such fine-grained reasoning about the performance of the vision are twofold. First, it provides planning and control modules with rich information that could be used for safe and optimal execution. Second, the extracted information can be used effectively to discover and categorize sources of errors for the vision system. While previous works on introspective vision systems ~\cite{zhang2014predicting, daftry2016introspective} output a single failure probability score for the whole input image, IVOA, to the best of our knowledge, is the first to digest the input image in detail and localize the potential sources of error and their type. 

We implement and test IVOA on a real-world dataset collected with a ground robot in both indoor and outdoor environments. We demonstrate IVOA's capability to accurately predict both false positive and false negative failure cases of two different stereo obstacle detection systems. We also show how IVOA's extracted information can be used to categorize sources of error for a vision system.

\vspace{2mm}
\section{RELATED WORK} \label{related_work}

\begin{figure*}[t]
  \centering 
  \includegraphics[width=1\linewidth,trim=0 0 0 -10,clip]{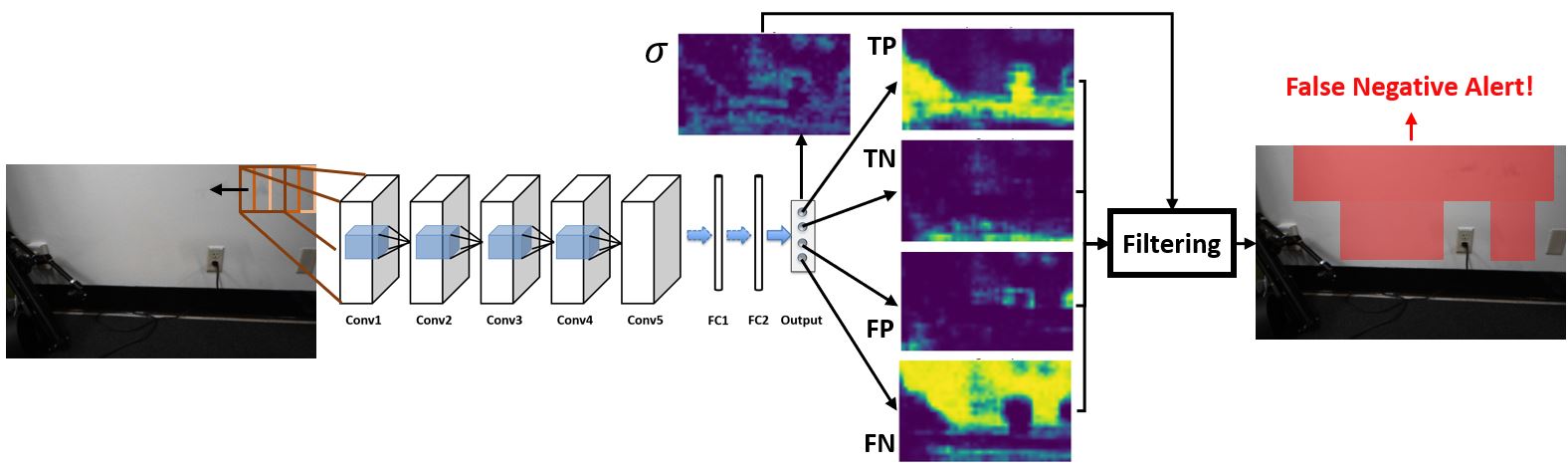}
  \caption{Pipeline of the introspection model. Image patches in the areas of interest on the input image are passed through a CNN to generate probability scores for different classes of failure. Potential areas of failure are detected  given the mean filtered probability scores and estimated uncertainty values.}
  \label{fig:NN_architecture}
  \vspace{-3mm}
\end{figure*}

In recent years, there has been a rise in research on introspective vision systems. One line of work tries to use the inherent uncertainty measure provided by the vision model. Grimmet et al.~\cite{grimmett2016introspective} look into probabilistic function approximators such as Gaussian processes as well as boot strapped classifiers that use the consensus of an ensemble of models as a measure of confidence. They evaluate the inherent uncertainty measure of these models by inspecting their changes when the model is exposed to new unseen data. Hu et al. \cite{hu2017introspective} tune the parameters of a localization algorithm by means of minimizing the inherent uncertainty measure of their perception model. Using only the underlying uncertainty of the perception model limits the introspective capacity of the system. A more rigorous approach is to train a second model, called the introspection model, to predict failure cases of the vision system. The introspection model does not need to know about the underlying details of the vision system. Instead it relies on the raw sensory input to predict the probability of failure of the vision system. Zhang et al.~\cite{zhang2014predicting} use labeled data and train a binary classifier, which given an input image predicts the success or failure of a vision system. They test their method for two different tasks of image classification and image segmentation. Daftry et al.~\cite{daftry2016introspective} train a convolutional neural network (CNN) that uses both still images and optical flow frames to predict the probability of failure for the navigation system of an actual UAV. A follow up work~\cite{saxena2017learning} trains an SVM classifier to choose the best recovery action when the vision system has high uncertainty. While all the above works predict only the overall reliability of the vision system given an input image, Ramanagopa et al.~\cite{ramanagopal2018failing} provides more detailed information, predicting and localizing false object detection instances on the input image for the use case of autonomous vehicles. They use stereo vision and leverage discrepancies between detected objects by each of the left and right cameras as cues for predicting failures. Their approach, however, is limited to predicting instances of false negatives and is specific to object detectors which do not suffice for safe obstacle avoidance.

Our work is similar to~\cite{daftry2016introspective} in that it seeks introspective vision for safe robot navigation. However, it goes beyond answering the question of whether the vision system may fail at a specific time step. Instead, IVOA predicts where in the input space the failure will happen and what type the failure will be. It should be noted that IVOA is distinct from works such as~\cite{ross2015online, hirose2018gonet, tai2016deep} that use only a black box model for the purpose of obstacle avoidance. It instead relies on a model-based stereo obstacle detection at the core and accompanies that with a black box model that provides predictions of failure cases of the former along with an uncertainty estimate of the predictions. We believe that this approach allows for robust and long-term deployment of robots in the real-world, where the robot will experience previously unseen environment.

\vspace{2mm}
\section{INTROSPECTIVE VISION SYSTEM} \label{introspection_model}

\mypara{Architecture}In~{\thisWork}'s architecture, the vision system consists of a perception and an introspection module. The perception module receives the raw sensory input and leverages its underlying model-based knowledge of the system, here stereo geometry, to provide the planning module with information about the surrounding of the robot. Unlike the perception module, the introspection is a black-box model and is responsible for assessing the reliability of the output of the perception given the same raw sensory input. In this work, we implement this vision system specifically for the purpose of obstacle avoidance for autonomous mobile robots. Our perception module is a stereo-vision based obstacle detector, that outputs an obstacle grid in front of the robot, where each cell in the grid is flagged as either obstacle free or occupied. The task of the introspection model is to predict, for each region on the input image, the probability of each of these four cases happening with regards to the perception module: 1-wrongly detecting an obstacle (false positive), 2-wrongly not detecting an obstacle (false negative), 3-correctly detecting an obstacle (true positive), 4-correctly detecting no obstacles (true negative).

\mypara{Perception Model} The perception model can be any stereo vision-based obstacle detection algorithm. The only requirement for the model is to be able to check the traversability of a point $(x, y, 0)$ in the reference frame of the robot, assuming it is in the field of view of the cameras. In this work we mainly use a version of the Joint Perception and Planning (JPP)~\cite{ghosh2017joint} algorithm. JPP utilizes a fast and computationally efficient method for detecting obstacles using stereo vision. Instead of creating a full dense reconstruction of the scene, it samples points of interest $(x, y, z)$ in the reference frame of the robot and projects them to the image planes of both cameras. Matching pairs of projections in the two image planes signal the existence of an object at the query point. We use JPP-C, which is JPP with the assumption of a convex world, i.e. obstacles are assumed to be on the ground (not hanging). We also test our approach on an implementation of dense stereo reconstruction with ELAS~\cite{geiger2010efficient}. This method creates a full 3D reconstruction of the scene via performing stereo matching for all pixels on the image and uses that for detection of obstacles.

\mypara{Introspection Model}We implement the introspection model as a multi-class classification convolutional neural network (CNN). We use the same layer architecture as the well known AlexNet~\cite{krizhevsky2012imagenet}, i.e. 5 convolution layers followed by 3 fully connected layers. The outputs of the last layer are passed through a softmax layer to provide normalized probability scores in the range $[0, 1]$. The model uses the image stream from only one of the cameras. Each $960 \times 600$ image obtained from the camera is sliced into overlapping $100 \times 100$ patches with a stride of $20$ pixels. Each patch is separately fed to the network and the outputs of network are scalar probability scores for each of the $4$ classes of false positive (FP), false negative (FN), true positive (TP), and true negative (TN). The output class probabilities of all patches are arranged in the original patches' configuration to form $49 \times 26$ heat maps of the probability of each class over all the input image. We want the introspection model to not only predict probability values for different classes of failure, but also to provide the degree of confidence it has in its prediction. We realize this by means of using two dropout layers before the first two fully connected layers during inference. Dropouts are mainly used for preventing overfitting in neural networks~\cite{srivastava2014dropout} during the training phase by randomly dropping units. Recent research, however, has shown that the same technique could be used during the inference phase to provide an estimate of uncertainty of the network~\cite{gal2016dropout}. We employ this technique in our network as following: each input image patch is passed through the network multiple times (we pick $20$), and at each pass different neurons are randomly dropped with a probability of $0.5$ at the dropout layers. The variance of the output of the network over these passes is taken as a measure of the introspection model's uncertainty for the given input image patch. In other terms, each input image is treated as a set of particles that pass through a stochastic model. The mean and variance of the output particles define the output of the model. The last stage of the introspection model is the post processing of the obtained probability scores. A mean filter is applied to the output probability heat maps of each class, and then at regions where the uncertainty is lower than some safety threshold classes with highest probability scores are announced as predictions. Fig.~\ref{fig:NN_architecture} shows the pipeline of the introspection model. 

\begin{figure}[t]
  \centering 
  \includegraphics[width=1\linewidth,trim=0 0 0 100,clip]{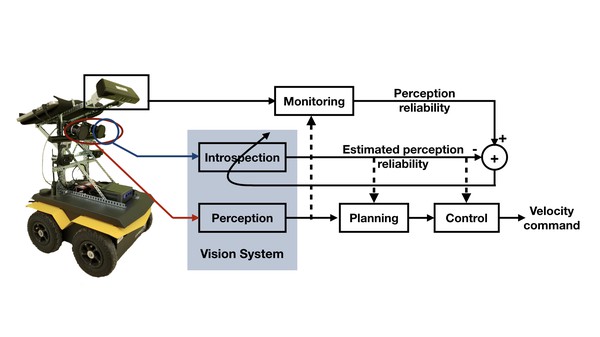}
  \caption{Training scheme for the introspection model. Depth readings by the Kinect sensor are used as the ground truth for automated generation of the training data.}
  \label{fig:introspection_model_training_diagram}
  \vspace{-3mm}
\end{figure}

\mypara{Training}We automate the training process for the introspection model by adding a high fidelity 3D depth sensor to the system. This sensor provides ground truth information for the monitoring module which in turn compares the depth sensor output with that of the perception module to generate labeled training data. Algorithm~\ref{alg:Train Data Generation} outlines the training data generation procedure. It should be noted that the depth sensor is only used for training. This training scheme helps reduce cost of large-scale robot deployments. Only a few of them need to be equipped with the costly monitoring depth sensor and the trained introspection model will be transferred to all robots. The ideal depth sensors to use in this system are 3D Lidars, which provide accurate depth readings of the surrounding environment upto long ranges and in various weather conditions. For a low-cost implementation of the system, however, we use a Kinect sensor to obtain ground truth depth readings. This limits us to training in indoor environments and outdoor environments only when there is not much sunlight as it interferes with the IR camera of the Kinect. Fig.~\ref{fig:introspection_model_training_diagram} illustrates the diagram of the navigation stack of the robot during training.

\begin{algorithm}[]
\caption{\textsc{Train Data Generation}}\label{alg:Train Data Generation}
\begin{algorithmic}[1]
\State $\textbf{Input:}$ Left camera image $I_l$, right camera image $I_r$, depth image $I_d$
\State $\textbf{Output:}$ Set of image patches $\{I_k\}_{k=1}^N$, set of ground truth labels $\{d_k\}_{k=1}^N$
\State $r\gets$ safety radius
\State $\mathbf{P}\gets \{(x', y', 0): X_{min} \le x' \le X_{max} \land
                                       Y_{min} \le y' \le Y_{max} \}$
\State $N\gets |\mathbf{P}|$
\For {$k\gets 1$ to $N$}
    \State $O_S\gets$ \textsc{IsObstacleFreeStereo}($I_l, I_r, p_k, r$)
    \State $O_M\gets$ \textsc{IsObstacleFreeMonitor}($I_d, p_k, r$)
    \State $(u, v)\gets$ \textsc{ProjectToLeftCam}($p_k$)
    \State $I_k\gets$ \textsc{ExtractPatch}($u, v$)
    \If {$O_M \land O_S$} ~ $d_k\gets TN$ 
    \ElsIf {$O_M \land \lnot O_S$} ~ $d_k \gets FP$
    \ElsIf {$\lnot O_M \land O_S$} ~ $d_k \gets FN$
    \Else ~ $d_k \gets TP$   
    \EndIf
\EndFor
\end{algorithmic}
\end{algorithm}

\section{EXPERIMENTAL RESULTS} \label{experimental_results}
\subsection{Evaluation Dataset}
We use the Clearpath Jackal, a mobile robot with a skid-steer drive system, for data collection. The robot is equipped with a stereo pair of Point Grey cameras that record $960 \times 600$ images at a rate of $30Hz$. Obstacle detection ground truth is provided by a Kinect depth sensor that is mounted on the robot and captures depth images at a rate of $30Hz$. The cameras and the Kinect are extrinsically calibrated with respect to each other. The robot is driven using a joystick and RGB and depth images are logged at full frame rate. The data is then processed offline: the depth images are converted to pointclouds and synchronized with the stereo camera images. For each set of synchronized images the perception module and the Kinect-based monitoring system are queried to determine whether a set of $(x, y, 0)$ points on a 2D grid in front of the robot and on the ground plane are obstacle free or not within a radius of $r = 10cm$. The corresponding pixel coordinates of the query points on the left camera's image plane along with the obstacle detection results are stored to form the dataset.
The indoor dataset spans multiple buildings with different types of tiling and carpet. The outdoor dataset is also collected on different surfaces such as asphalt, concrete, and tile in both dry and wet conditions. The total dataset of more than $1.4km$ traversed by the robot includes about $6$ million extracted image patches and $120k$ full image frames. The data is split into train and test datasets, each composed of separate full robot deployment sessions, such that both train and test sets include data from all types of terrain.

\subsection{Evaluation Metric}
\begin{figure}[t]
  \centering 
  \includegraphics[width=1\linewidth,trim=0 0 0 0,clip]{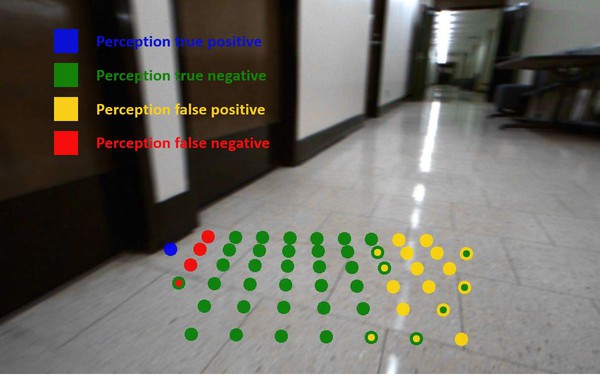}
  \caption{Comparing the ground truth label of the points on a 2D grid with the predicted labels by the introspection model. For each point on the grid, the inner circle's color represents the prediction of the introspection model and the outer one indicates the ground truth result of the perception system.}
  \label{fig:evaluation_metric}
  \vspace{-3mm}
\end{figure}

The performance of the introspection model is evaluated based on its ability to predict the behavior of the perception model. For each image the introspection model is queried with the same points on the image, for which we have the  prediction result of the perception system as one of the four classes of FP, FN, TP, and TN. The accuracy of the model in predicting each of these classes is assessed as a measure of its performance. Fig.~\ref{fig:evaluation_metric} denotes an example of comparing the output of the introspection model against the ground truth.

\subsection{Model Accuracy Results}
We train the introspection model on the train dataset and then test the model separately on the indoor and outdoor portions of the test dataset.
Please note that for the rest of the paper until the end of Section~\ref{sec:error_source}, the reported results correspond to {\thisWork} using JPP-C as the perception model. In Section~\ref{sec:perc_model_comparison} we show results of {\thisWork} trained on ELAS. 

\begin{figure}[t]
  \centering 
  \includegraphics[width=0.9\linewidth,trim=0 0 0 0,clip]{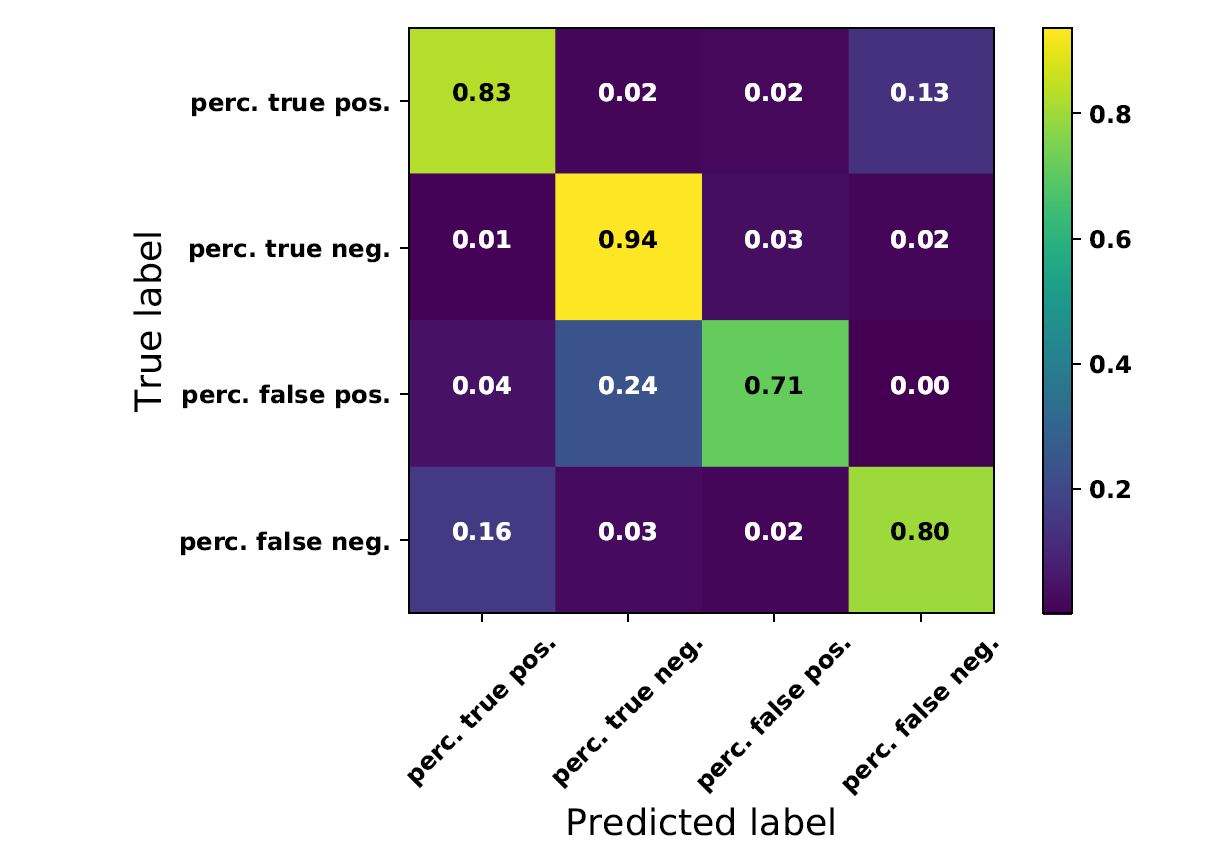}
  \caption{Prediction results of the introspection model on the indoor dataset.}
  \label{fig:agg_results_indoor}
  \vspace{-3mm}
\end{figure}

\begin{figure}[t]
  \centering 
  \includegraphics[width=0.9\linewidth,trim=0 0 0 0,clip]{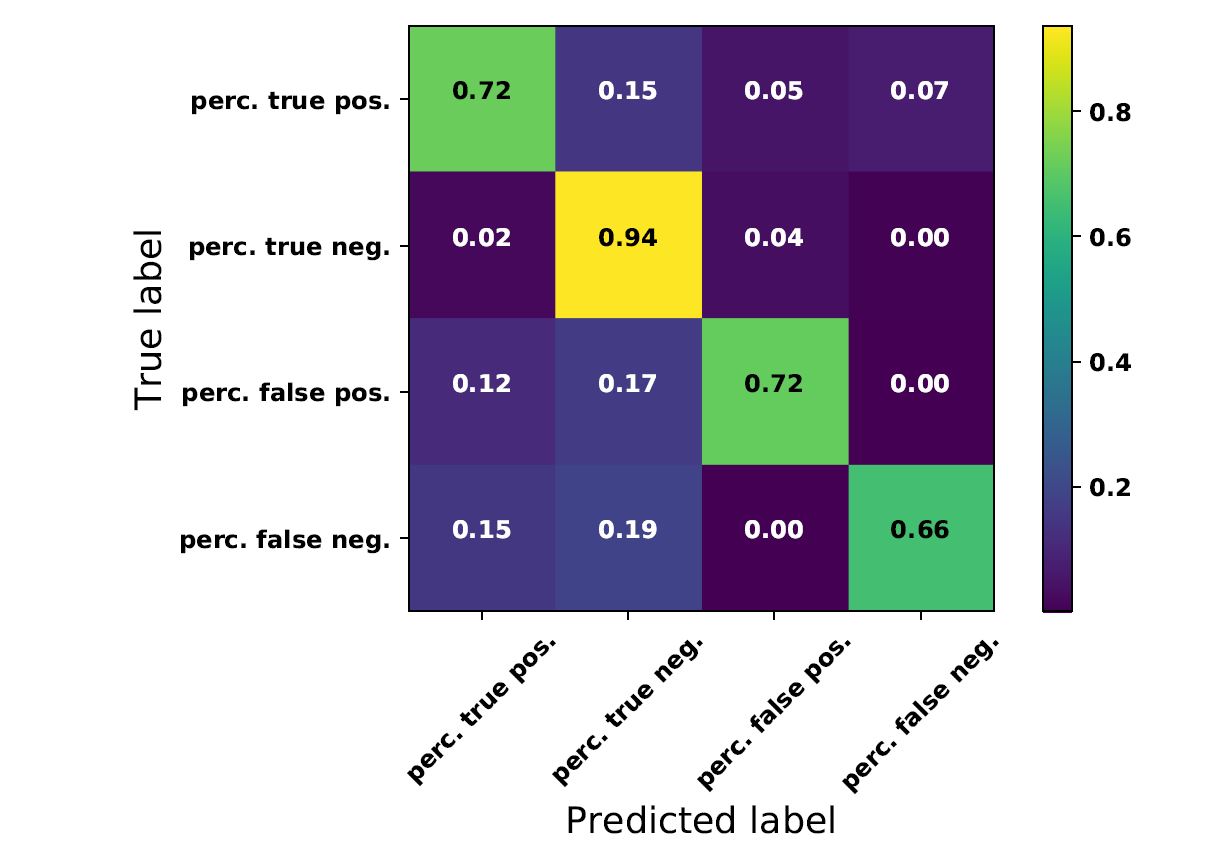}
  \caption{Prediction results of the introspection model on the outdoor dataset.}
  \label{fig:agg_results_outdoor}
  \vspace{-3mm}
\end{figure}

\begin{figure}[t]
  \centering 
  \includegraphics[width=1\linewidth,trim=0 0 0 0,clip]{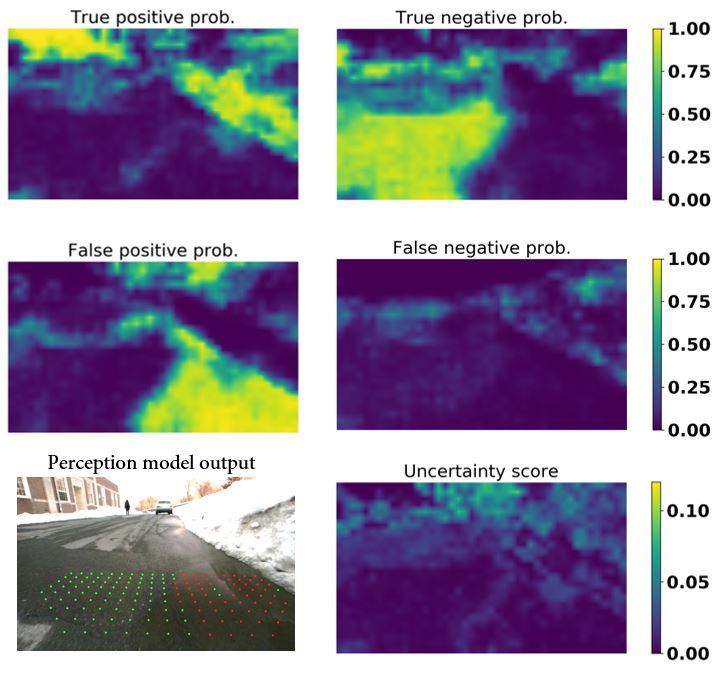}
  \caption{Example output of the introspection model. The bottom left image shows the estimated obstacle grid in front of the robot by the perception model, where the red and green dots denote the detected occupied and obstacle free cells respectively. The introspection model correctly predicts water reflections to cause false positives for the perception model (middle left image).}
  \label{fig:detailed_introspection_output}
  \vspace{-3mm}
\end{figure}

 In this section, we present the results with the uncertainty-based filtering of the introspection model turned off, i.e. the model classifies all data points as one of the four classes even if the uncertainty level is high. We analyze the effect of model uncertainty in the next section. The results are demonstrated in Fig.~\ref{fig:agg_results_indoor},\ref{fig:agg_results_outdoor}. The introspection model is able to catch a significant portion of the failure cases and predict their type correctly for both indoor and outdoor datasets. It is interesting to note that even in cases when the introspection model is not able to predict a failure, it still correctly detects the existence of an obstacle. Fig.~\ref{fig:detailed_introspection_output} shows the detailed outputs of the introspection model for an example input image.

\subsection{Effect of Model Uncertainty} \label{uncertainty_results} 
\begin{figure}[t]
  \begin{subfigure}[b]{0.22\textwidth}
    \includegraphics[width=40mm,trim=0 0 0 0,clip]{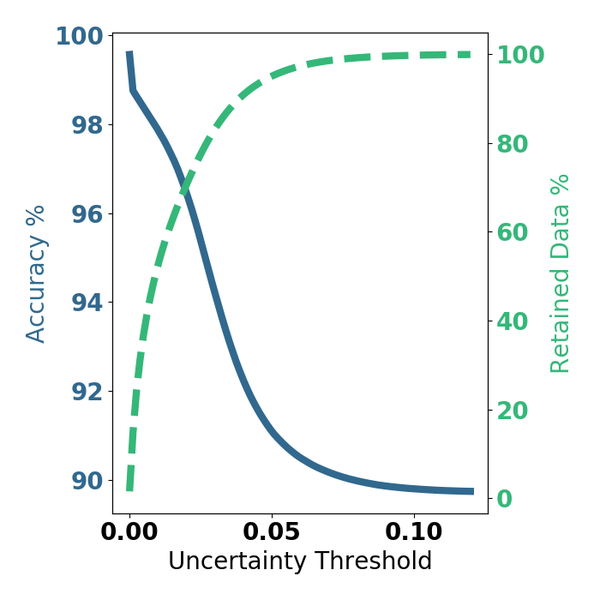}
  	\caption{}
  	\label{accuracy_uncertainty_plot}
  \end{subfigure}
  \begin{subfigure}[b]{0.22\textwidth}
    \includegraphics[width=40mm,trim=0 0 0 0,clip]{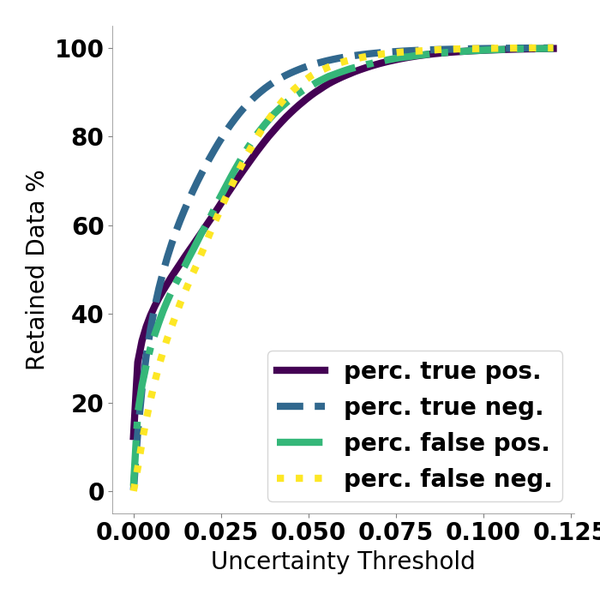}
  	\caption{} 
  	\label{retained_data_uncertainty_plot} 
  \end{subfigure}
  \caption{The effect of introspection model's uncertainty threshold. (a) The solid graph shows the total accuracy of the introspection model over all classes after retaining only part of the data for which the model uncertainty is less than the threshold. The dashed graph shows the percentage of retained data for different uncertainty thresholds. (b) The percentage of retained data vs. uncertainty threshold for different classes of data points.}
  \label{fig:uncertainty_thresh}
  \vspace{-3mm}
\end{figure}

As explained in section~\ref{introspection_model}, our proposed introspection model provides an estimate of its inherent uncertainty. In this section, we analyze the importance of the uncertainty measure in the reliability and performance of the system. We run the introspection model on the whole test dataset. We then sort the data based on the introspection model's uncertainty score in ascending order. We start removing data points from the bottom of the list, whose uncertainty score is higher than an uncertainty threshold value. The accuracy of the introspection model is then calculated on the retained data points as the mean of the prediction accuracy values for each classe. Fig.~\ref{accuracy_uncertainty_plot} illustrates that the accuracy increases monotonically with decrease in the uncertainty threshold. At the point, when still $70\%$ of the data is retained, it reaches an accuracy of more than $96\%$ with a $6\%$ improvement compared to not using the uncertainty measure. Fig.~\ref{retained_data_uncertainty_plot} also shows the percentage of the retained data for each class and over the same range of uncertainty thresholds. As can be seen in the figure, the rate of dropping data is roughly the same for all $4$ classes. The results prove the correctness of the estimated uncertainty measure, in that it is inversely correlated to   the accuracy of model. It should be noted that such uncertainty measure is of paramount importance especially for a failure detection system that is based on a black-box model. Deep learning models are prone to making false predictions when exposed to unseen and totally new inputs. Using an uncertainty estimation, however, reduces such failures and makes neural networks suitable for use in real-world applications such as robotics.

\subsection{Categorizing Sources of Error} \label{sec:error_source}
As mentioned earlier, one of the motivations of {\thisWork} from performing a fine-grained failure detection is to behave as an assisting tool for debugging of vision systems. In order to test this hypothesis, we try clustering the detected instances of failure. From all instances of false positive and false negative, detected by the introspection model and on the test dataset, we pick the top $50\%$ in terms of the confidence of the predictions. Then for each corresponding image patch $\mathbf{I}_i$, the normalized output of the second fully connected layer of the introspection model $\mathbf{x}_i \in \mathbb{R}^{256}$ is extracted as an embedding. 

\begin{figure*}[t]
  \centering 
  \includegraphics[width=1\linewidth,trim=0 0 0 0,clip]{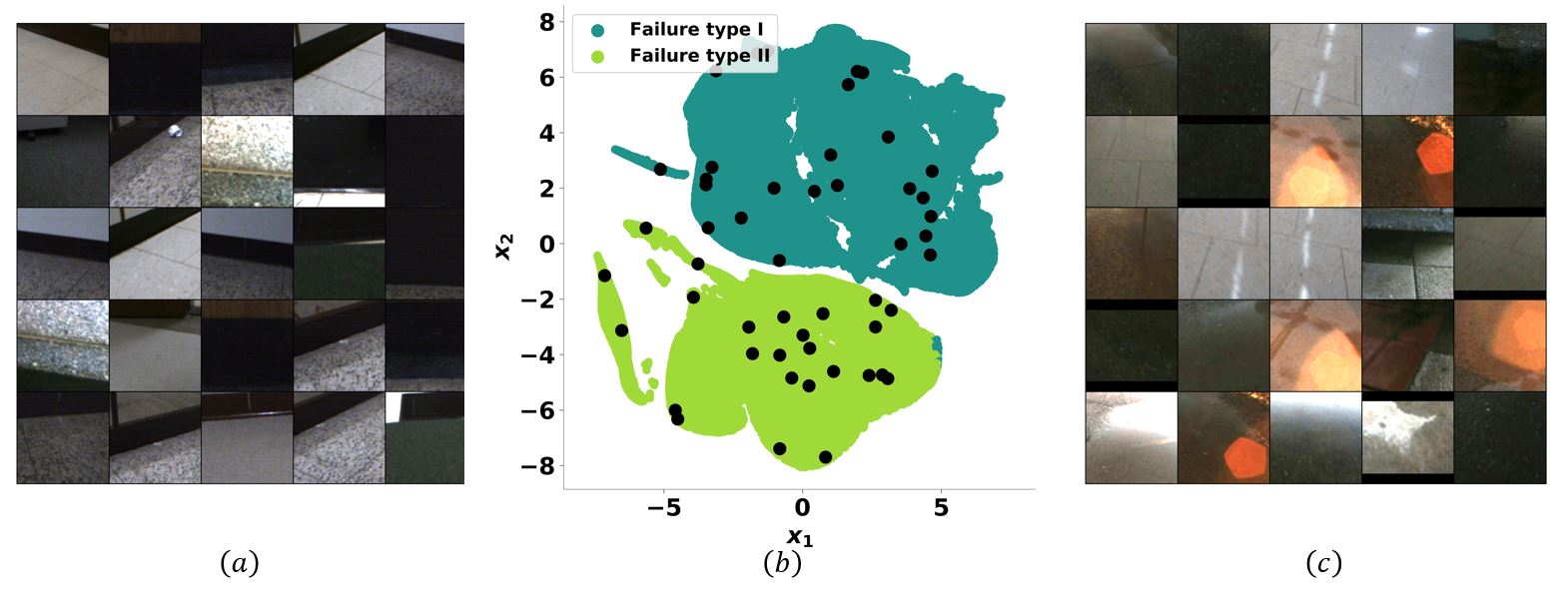}
  \caption{Clustering predicted sources of error. b) Two extracted clusters of perception failures in the embedding space projected down to a $2D$ space via dimensionality reduction. a) Randomly sampled image patches from failure type II cluster, shown as black dots on (b). c) Image patches randomly sampled from failure type I cluster. {\thisWork} detects reflection and glare (type I) and the dark stripes at the bottom of the walls (type II) to be the most dominant sources of error for the perception model under test.}
  \label{fig:failure_case_clustering_wide}
  \vspace{-3mm}
\end{figure*}

In order to decide on the number of clusters, we first visualize a 2D representation of the data. PCA is performed to reduce the dimension of embeddings by a factor of $10$, and then t-SNE~\cite{maaten2008visualizing}, a nonlinear dimensionality reduction approach well-suited for visualization of high dimensional data, is applied to obtain a 2D representation of the samples. Based on the result of the visualization, a cluster number of $2$ is chosen and k-means clustering is applied to the data in the original embedding space. Fig.~\ref{fig:failure_case_clustering_wide} illustrates the resultant clusters projected down to the 2D space as well as sampled image patches from each cluster. The result shows that {\thisWork} the dark edges at the bottom of the walls and reflection/glare to be the most dominant sources of error for the perception model under test.

\subsection{Adaptability to Different Perception Systems}\label{sec:perc_model_comparison}
\begin{figure}[t]
  \centering 
  \includegraphics[width=1\linewidth,trim=0 0 0 0,clip]{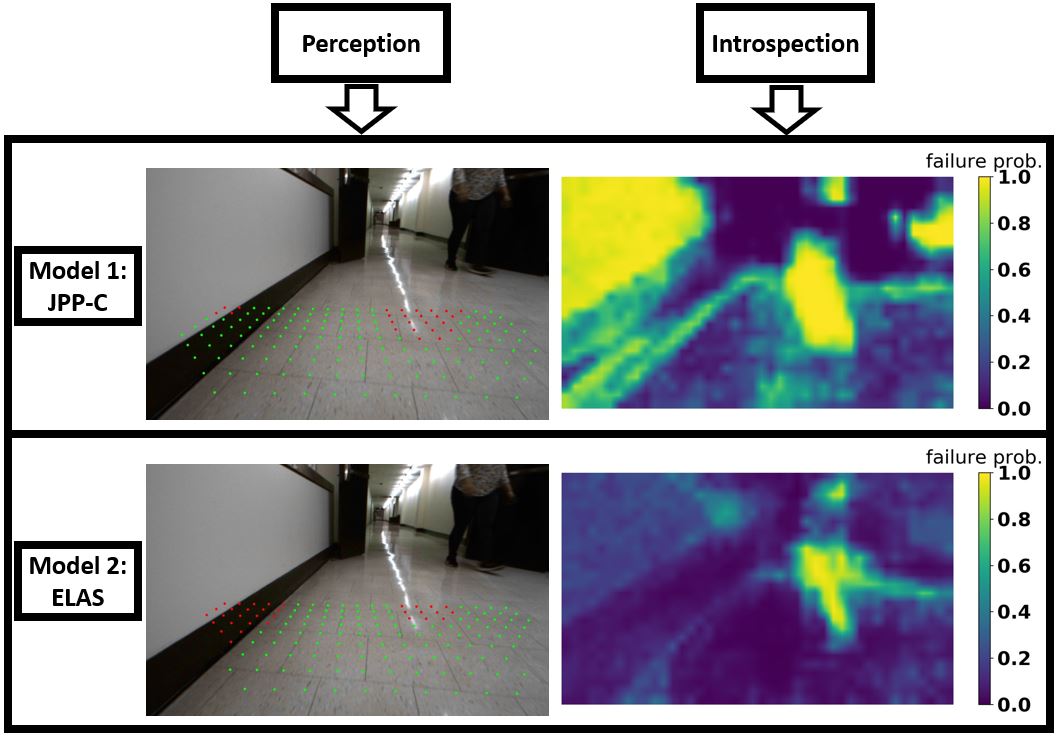}
  \caption{Introspection model's prediction for different perception models. On the left column, the produced obstacle grids by each of the two perception models are visualized on the input image. The green and red dots represent the detected obstacle free and occupied cells respectively. On the right, output of the introspection model is shown as the total probability of failure $\Pr(FP) + \Pr(FN)$ for each perception model. The introspection model correctly predicts the reflection to cause false positives for both models and the wall to cause false negatives for only one of them.}
  \label{fig:new_perception_model}
  \vspace{-3mm}
\end{figure}

Our proposed architecture of the introspective vision system as explained in section~\ref{introspection_model} is agnostic to the perception model. Any obstacle detection system can be used in place of JPP-C, and the training and inference of the system will remain intact. In order to test this feature, we trained the introspection model for an obstacle detection system based on stereo dense reconstruction of the scene using the ELAS~\cite{geiger2010efficient} stereo matching technique. The resulting introspection model was able to learn failure cases of the new perception model. Fig.~\ref{fig:new_perception_model} compares the output of the two different perception systems alongside the introspection model's prediction of their performance for the same scene. Both JPP-C and ELAS wrongly detect the reflection on the tile as an obstacle (hole in the ground). Also, JPP-C fails to detect the texture-less wall, while ELAS is able to correctly detect it. As shown in the figure, the introspection model correctly predicts the behavior of both models.

\section{CONCLUSION} \label{conclusion}

In this paper, we introduced {\thisWork}: an architecture for competency-aware stereo vision-based obstacle avoidance systems capable of predicting their failures, while distinguishing between false positive and false negative instances. We demonstrate {\thisWork}'s ability to accurately predict failures of the vision on a real-world dataset in both indoor and outdoor environments. As future work, we would like to integrate {\thisWork} with planning and control to leverage its detailed estimate of the reliability of vision for safe and optimal navigation of mobile robots.

\bibliographystyle{IEEEtran}
\bibliography{IEEEabrv,bibliography}

\end{document}